\documentclass{article}

\usepackage[letterpaper,top=2cm,bottom=2cm,left=3cm,right=3cm,marginparwidth=1.75cm]{geometry}
\usepackage[english]{babel}
\usepackage{graphicx}
\usepackage[colorlinks=true, allcolors=blue]{hyperref}

\usepackage{authblk} 

\usepackage[sort,numbers]{natbib}
\usepackage{amsfonts}
\usepackage{amsmath}
\usepackage{amsthm}
\usepackage{array}
\usepackage{enumitem}

\newcommand{\norm}[1]{\left\lVert#1\right\rVert}

\newtheorem*{problem*}{Problem}

\newtheorem*{question*}{Question}
\newtheorem{question}{Question}

\author[1]{Lei Zhang}
\author[2]{Heung-Yeung Shum}
\affil[1,2]{International Digital Economy Academy, Shenzhen 518048, China}
{
    \makeatletter
    \renewcommand\AB@affilsepx{: \protect\Affilfont}
    \makeatother

    \affil[ ]{Email}

    \makeatletter
    \renewcommand\AB@affilsepx{, \protect\Affilfont}
    \makeatother

    \affil[1]{leizhang@idea.edu.cn}
    \affil[2]{hshum@idea.edu.cn}
}

\title{Statistical Foundation Behind Machine Learning and \\Its Impact on Computer Vision}
\date{}

\begin{document}
\maketitle

\begin{abstract}
This paper revisits the principle of uniform convergence in statistical learning, discusses how it acts as the foundation behind machine learning, and 
attempts to gain a better understanding of the essential problem that current deep learning algorithms are solving. Using computer vision as an example domain in machine learning, the discussion shows that recent research trends in leveraging increasingly large-scale data to perform pre-training for representation learning are largely to reduce the discrepancy between a practically tractable empirical loss and its ultimately desired but intractable expected loss. Furthermore, this paper suggests a few future research directions, predicts the continued increase of data, and argues that more fundamental research is needed on robustness, interpretability, and reasoning capabilities of machine learning by incorporating structure and knowledge.
\end{abstract}

\section{Background}

Since 2012, the breakthrough in deep learning has led to remarkable progress in computer vision~\citep{lecun2015deep, bengio2021deep}. For example, for image classification, the top-1 accuracy of 1000-class classification on ImageNet has been dramatically improved from 50.9\% before 2012 to 91.0\% in 2022~\citep{paperwithcode2022imagenet, yu2022coca}. For object detection, the mAP (mean average precision) was only 53.3\%~\citep{girshick2014rich} in 2014 on Pascal VOC 2012 for 20 object categories and has been improved to 68.17\%~\citep{liu20201st} in 2019 on Open Images for 500 object categories, which are significantly more than 20 categories in VOC. The algorithms developed for image classification and object detection have become fundamental building blocks empowering many other vision problems. See~\citep{voulodimos2018deep} for a more comprehensive survey of deep learning for computer vision.

However, looking at real applications that need computer vision techniques, we still see a big gap between the current state-of-the-art (SOTA) and the desired performance in real scenarios. Even for face recognition, which has been widely used in real applications, its algorithms still suffer from the bias issue and are vulnerable to adversarial attacks. For object detection, as the generic detection performance is only as good as mAP 68.17\% (for 500 categories on Open Images), its application has to be narrowed to a handful number of object categories to make it practically usable, e.g. face detection and vehicle detection, or constrained to non mission-critical problems, e.g. digital asset management.

Natural language processing (NLP) has witnessed a similar phenomenal progress but with even bigger challenges. Unlike computer vision, which is normally regarded as a sensing problem, language understanding often needs to deal with knowledge and reasoning, which lead to many critiques arguing its lack of theoretical foundation in addressing such problems. Such critiques are not only for the domain of NLP, but also for the whole area of artificial intelligence (AI). For example, \citet{jordan2019artificial} pointed out that the developments now being called AI arose mostly in the engineering fields associated with low-level pattern recognition and movement control, as well as in the field of statistics, and their underlying systems do not involve high-level reasoning or thought. \citet{pearl2018build} amounted all the impressive achievements of deep learning to just curve fitting and expected that causal reasoning could provide machines with human-level intelligence. \citet{marcus2018deep} regarded deep learning a brute force approach, which works less well when there are limited amounts of training data, or when the test set differs importantly from the training set, and proposed a hybrid, knowledge-driven, reasoning-based approach, centered around cognitive models for a more robust AI~\citep{marcus2020next}. \citet{zhang2020towards} suggested that to address the deficiencies of the current AI technologies, it is imperative to combine the two competing paradigms of AI development since 1956, i.e., symbolism and connectionism, to develop a new, explainable, and robust AI theory and method. 

Despite all the challenges and critiques, the recent progress in deep learning is undeniable. However, there are few discussions \emph{why} deep learning is effective in some fields but fails in others from the perspective of statistical learning. One often attributes the success of deep learning to three factors: algorithm, big data, and computing power~\citep{zhang2020towards}, and call current deep learning-based algorithms data-driven approaches. Indeed, in the past few years, the scale of the dataset, the number of model parameters, and the number of GPUs used for training a language or vision foundation model have been tremendously increased~\citep{yuan2021florence, yu2022coca, brown2020language, wudao2021}. As a result, records are constantly refreshed on various benchmarks, e.g. ImageNet~\citep{deng2009imagenet}, COCO~\citep{lin2014microsoft}, WMT2014~\citep{bojar2014findings}, GLUE~\cite{wang-etal-2018-glue}, etc. To understand how the factors of algorithm, big data, and computing power are connected and where they succeed or fail, it is worthwhile to revisit the statistical foundation behind machine learning, gain a better understanding of what essential problem the deep learning algorithms are solving, and develop solutions that can possibly address the shortcomings of the current approaches.

The purpose of this paper is not to propose any new concept, but to show the fundamental connection between classical statistical learning and modern machine learning (particularly in the form of deep learning), which is often overlooked and less discussed in research papers in computer vision. The discussion about the statistical foundation behind machine learning is general to any application domains. But the paper uses computer vision as an example domain to make the discussion more accessible to researchers and practitioners in related fields. Through the discussion, we hope one can gain a better understanding of the current state of the art: what works, what does not, what is the limitation of current data-driven approaches, what we should improve, etc. In the end, we will suggest a few future research directions, predict the continued increase of data, and argue that more fundamental research is needed to address the weakness of data-driven approaches by incorporating structure and knowledge.

\section{Statistical Foundation Behind Machine Learning}

\subsection{What is machine learning?}\label{sec:what_is_ml}
Machine learning, a term popularized by \citet{samuel1959some}, is generally viewed as a sub-field of artificial intelligence. Over the past decades, many machine learning algorithms have been developed to empower computers to learn from data and make predictions or decisions without being explicitly programmed. Deep learning refers to machine learning algorithms that employ deep neural networks, and has led to great progress in many domains, including, but not limited to, computer vision, natural language processing, speech recognition, game, and robotics.

Generally speaking, machine learning can be regarded as solving the following problem.

\begin{problem*}
Assuming that we have seen $N$ data pairs $\{(x_i,y_i)\}_{i=1}^N$, where 
$x_i\in\mathcal{X}$ is an observed data sample and $y_i\in\mathcal{Y}$ is its associated label, can we predict something about the $(N+1)$-th data sample $x_{N+1}$? 
\end{problem*}

This is also known as inductive reasoning which has been widely used in the development of science. Science is aiming at understanding the causal relationship from observed data so that important laws such as Newton's law of universal gravitation can be discovered. Machine learning is trying to directly learn the relationship from data, though mostly being correlated instead of causal.

Mathematically, machine learning is to learn a mapping function $f:\mathcal{X}\rightarrow\mathcal{Y}$ from a given set of functions parameterized by $\alpha\in\Lambda$, which can be formulated as $y=f(x|\alpha)$. We also know that $(x,y)$ follows a fixed (but unknown) probability distribution $F(x,y)$, from which we can sample data. For example, predicting whether an image has a face or not is a binary classification problem. Generating a sentence to describe an image is an image-to-text caption generation problem.

To make the problem tractable, a machine learning approach typically draws $N$ independent and identically distributed (i.i.d.) samples $\{(x_i,y_i)\}_{i=1}^N$ from $F(x,y)$ and uses the sampled data as a \textit{proxy} to define a learning problem and then finds its optimal parameter $\alpha$. The intuition is, if $y=f(x|\alpha)$ \textbf{explains} $\{(x_i,y_i)\}_{i=1}^N$ well, we \textbf{hope} it also works for $x_{N+1}$.

Here, ``\textbf{explain}'' and ``\textbf{hope}'' are intuitive terms. For example, in many computer vision problems, one often trains a model on a finite set of training images. One of the criteria for a good model is that its prediction on any training image matches with its ground truth label. That is, the trained model explains\footnote{Note that explainability is a different term in the literature. It is concerned that if the model prediction is a white-box process which provides results that are understandable for experts in the domain~\citep{wiki2022explainability}. For example, a model that lacks explainability can make unreasonable-to-human wrong predictions on adversarial inputs, for example, predicting a panda image with an imperceptibly small perturbation as ``gibbon''~\citep{goodfellow2015explaining}.} well the training data. But the goal of machine learning is to make prediction on unseen data, which are normally infinite in real applications. The problem of ``hope'' is if the trained model can make good predictions on unseen data. It is crucial to understand the statistical conditions when a model trained on finite data generalizes on infinite unseen data.

Next we will discuss what means ``\textbf{explain}'' and what means ``\textbf{hope}'' in more details.

\textbf{What means ``explain''?} To check if $y=f(x|\alpha)$ \emph{explains} $\{(x_i,y_i)\}_{i=1}^N$ well, we can define a loss function $Q(f(x|\alpha),y)$ to measure the difference between a predicted value and its desired value. The overall loss is defined by averaging the differences over $N$ data pairs:

\begin{equation}\label{eq:loss_emp}
L_{emp}(\alpha)=\frac{1}{N}\sum_{i=1}^{N}Q(f(x_i|\alpha),y_i).
\end{equation}

Note that this loss function depends on the sampled $N$ data pairs, thus is called \textit{empirical} loss. Compared with the general inductive reasoning principle utilized in the development process of science, Eq. \eqref{eq:loss_emp} offers a practical solution that can be solved by a computer rather than a genius scientist.

Eq. \eqref{eq:loss_emp} is very general. It helps to explain why three factors \textit{algorithm}, \textit{big data}, and \textit{computing power} are essential for the success of machine learning and AI.

\begin{itemize}
    \item \textbf{Algorithm} - designing a proper mapping function $f(x|\alpha)$, parameterized by $\alpha\in\Lambda$
    \item \textbf{Big data} - defining an empirical loss $L_{emp}(\alpha)$ to tell which function is better
    \item \textbf{Computing power} - solving the optimization problem  $\hat{\alpha}=\operatorname*{arg\,min}_{\alpha\in\Lambda} L_{emp}(\alpha)$
\end{itemize}

\subsubsection{Mapping function \texorpdfstring{$f(x|\alpha)$}{Lg}}
The mapping function reflects our understanding to the learning problem, i.e. how the input variable $x$ is mapped to the output variable $y$. Prior to deep learning, linear function is most widely used and studied:
$$
f(x|\alpha)=W x +b,
$$
where $\alpha=\{W, b\}$.

Deep learning provides a more principled way for designing a nonlinear mapping function in a layer-by-layer manner:
$$
f(x|\alpha)=g\Bigg(h\bigg(k\Big(r\big(q\left (...  \right )  \big)  \Big)  \bigg)  \Bigg).
$$

In the past decade, the research community has developed and open-sourced many easy-to-use building blocks such as fully connected layer, convolutional layer, and Transformer encoder and decoder layers, which significantly helped the design of mapping functions. 

\subsubsection{Loss function \texorpdfstring{$Q(f(x|\alpha), y)$}{Lg}}\label{subsec:loss}
The loss function form in Eq. \eqref{eq:loss_emp} covers most machine learning problems. Below we list a few examples for supervised learning:

\begin{enumerate}
\item Binary classification:

$$
Q(f(x|\alpha),y)=
\begin{cases}
0,  & \text{if $f(x|\alpha)=y$} \\
1,  & \text{if $f(x|\alpha)\neq y$},
\end{cases}    
$$
where $y\in\{0,1\}$ is a discrete binary value and $f$ is a binary prediction function.

\item K-class classification (cross entropy):
$$
Q(f(x|\alpha),y)=\sum_{k=1}^K y_k \log f_k(x|\alpha),
$$
where $y\in[0,1]^K$ and $\sum_{k=1}^K y_k=1$ is a K-dimensional probability distribution and $f$ is a probability prediction function.

\item Regression (L2 loss):
$$
Q(f(x|\alpha),y)=\norm{y-f(x|\alpha))}_2^2,
$$
where $y\in \mathbb{R}^n$ is an $n$-dimensional real-valued vector and $f$ is a real-valued vector prediction function.
\end{enumerate}

The loss function $Q(f(x|\alpha),y)$ also applies to unsupervised learning, where the supervision $y$ is simply the input $x$ itself, i.e. $y=x$. Below is an example for L2-based loss, while many other forms can also be introduced.
$$
Q(f(x|\alpha),x)=\norm{x-f(x|\alpha))}_2^2.
$$

Such a loss is essentially to perform a reconstruction task, with interest to discover intrinsic structures from data for more robust representation learning. More discussions please see Question \ref{question:unsupervised} in Sec. \ref{sec:discussion}.

\subsection{Statistical Foundation}\label{sec:foundation}
The empirical loss $L_{emp}$ only deals with $N$ seen data, aiming to learn a mapping function $y=f(x|\hat{\alpha})$ that can \emph{explain} $\{(x_i,y_i)\}_{i=1}^N$ well if $\hat{\alpha}$ minimizes $L_{emp}$. However, optimizing $L_{emp}$ is no more than a curve fitting problem. How can we \emph{hope} the learned mapping function also works for $x_{N+1}$?

One may argue that we can use a test set $\{(x_i,y_i)\}_{i=1}^M$ to \emph{measure} the performance of the learned function, which is a widely used practice in machine learning. That is, after we find the optimal parameter $\hat{\alpha}$ that minimizes $L_{emp}$, we can test the loss on $\{(x_i,y_i)\}_{i=1}^M$.

\begin{equation}\label{eq:loss_test}
L_{test}(\hat{\alpha})=\frac{1}{M}\sum_{i=1}^{M}Q(f(x_i|\hat{\alpha}),y_i).
\end{equation}

Note that the test set is also sampled from the same probability distribution $F(x,y)$ and subjects to sampling bias. When the size of the test set is limited and often smaller than the training set as in most applications, $L_{test}$ can only be used as an estimator and will inevitably lead to an estimation variance. There is no theoretical guarantee to ensure that the optimal parameter $\hat{\alpha}$ also optimizes $L_{test}$.

\begin{figure*}[ht]
    \centering
    \includegraphics[width=0.8\textwidth]{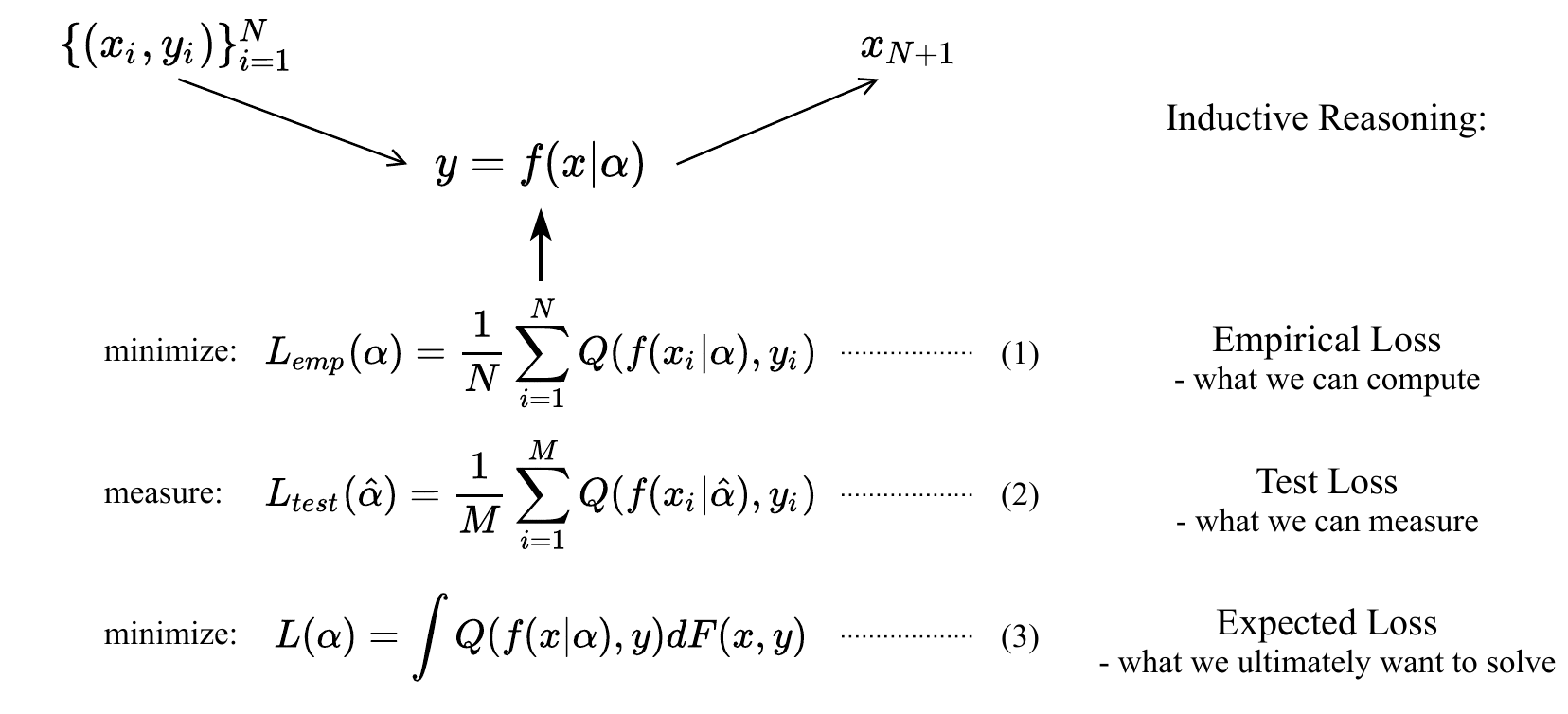}
    \caption{Machine learning is to learn a mapping function from observed data in order to make predictions for unseen data. In practice, one normally minimizes an empirical loss defined over observed data and measures the performance of the estimated parameter $\hat{\alpha}$ on another i.i.d test data set. But it is worth noting that the ultimate goal is to minimize the expected loss. Statistical learning is concerned that if the two objectives defined in Eq. (1) and Eq. (3) are consistent.}
    \label{fig:machine_learning}
\end{figure*}

\textbf{What means ``hope''?} To \emph{hope} the learned mapping function also works for $x_{N+1}$, we need to review our learning problem again. As shown in the beginning of Sec. \ref{sec:what_is_ml}, machine learning solves the problem of learning a mapping function $y=f(x|\alpha)$ from the sampled data set $\{(x_i,y_i)\}_{i=1}^N$, which follows a fixed (but unknown) probability distribution $F(x,y)$. However, note that our true motivation is finding a right mapping function $y=f(x|\alpha)$ that works for unseen data. The sampled data set is only a proxy that helps us define an empirical loss $L_{emp}$ and makes the learning problem tractable.

The true \emph{expected} loss function should be
\begin{equation} \label{eq:loss_expected}
    L(\alpha)=\int Q(f(x|\alpha), y)d F(x,y).
\end{equation}

As shown in Fig. \ref{fig:machine_learning}, statistical learning is concerned with the following question: is minimizing the empirical loss $L_{emp}$ in Eq. \eqref{eq:loss_emp}) consistent with minimizing the expected loss $L$ in Eq. \eqref{eq:loss_expected}? That is, if and when the following convergence takes place for any positive $\epsilon$,

\begin{equation}\label{eq:uniform_convergence}
P \Bigg\{\sup_{\alpha\in\Lambda}\Bigg|\int Q(z,\alpha)d F(z) - \frac{1}{N}\sum_{i=1}^{N}Q(z_i,\alpha)  \Bigg|>\epsilon \Bigg\}
\underset{N\rightarrow\infty}{\xrightarrow{\hspace*{0.8cm}}}0,
\end{equation}
where for simplicity we have replaced $Q(f(x|\alpha), y)$ with $Q(z,\alpha)$ and $F(x,y)$ with $F(z)$. Note that $\{z_i\}_{i=1}^N$ are random variables sampled from $F(z)$. The probability is computed over all possible sampled random variables for any given $N$.

This is essentially the generalization problem for the inductive reasoning principle, which has been thoroughly studied in statistical learning \citep{vapnik1998}. The study has led to the generalization of the Law of Large Numbers in functional space.

When $\Lambda$ has only one (or a finite number of) element, i.e. only one choice of $\alpha$, Eq. \eqref{eq:uniform_convergence} is reduced to the Law of Large Numbers, which states that the average of the observation values converges to the expected value when the number of trials $N$ goes to infinity.

This special case can be regarded as model evaluation after the learning process is done and an optimal parameter has been found. But even in this case, to evaluate the real \emph{expected} performance, a sufficiently large number of test examples is needed. Evaluation is very difficult in many applications as it requires intensive efforts to develop a comprehensive evaluation data set. 

When $\Lambda$ has an infinite number of elements, Eq. \eqref{eq:uniform_convergence} does not necessarily converges. Its convergence is the generalization of the Law of Large Numbers in functional space. In statistical learning, the necessary and sufficient condition that ensures this convergence was proved in 1980s \citep{vapnik1998}. The condition is mainly to bound the capacity of the function set $\{f(x|\alpha)| \alpha\in\Lambda\}$. Otherwise, if the function set had an unlimited capacity, it would memorize any given training set and lead to an empirical loss of 0 but cannot generalize to any unseen data.

Eq. \eqref{eq:uniform_convergence}, which is called uniform convergence, is profound in statistical learning. From Eq. \eqref{eq:uniform_convergence} and its necessary and sufficient condition about the function capacity, we can see three essential factors for effective learning.

\begin{itemize}
    \item \textbf{Big data} -- Large-scale data not only helps define an approximation ( empirical loss) to a target problem, but also is a necessary condition to ensure the asymptotic convergence of an empirical loss to its expected loss. A smaller scale data set will inevitably lead to the variance error due to the randomness of the sampled data. Moreover, data is never sufficient in real applications. Take face/non-face classification on an input image of $20\times20$ with 16 level gray values as a simple example. The total number of all possible samples would be $16^{20\times20}\approx 4.5\times10^{481}$, which is simply impossible to enumerate. To mitigate this issue, some priors over the mapping function $f(x|\alpha)$ or data manifold have been introduced for better generalization. For example, some priors are function smoothness, low-dimensional data manifold, multiple and hierarchical explanatory factors, etc. See \citep{bengio2013representation} for more discussions.
    
    \item \textbf{i.i.d assumption} -- Uniform convergence assumes the sampled data are independent and identically distributed. The \textit{independent} assumption makes it possible to decouple the joint distribution of $p(z_1, z_2, ..., z_N)$ into the multiplication of distributions over individual samples $\prod_{i=1}^N p(z_i)$ and then convert it to the sum of log probabilities $\sum_{i=1}^N \log p(z_i)$. To see what it means in deep learning, empirical loss is typically defined as an additive loss (due to the i.i.d. assumption) over every training sample and then stochastic gradient descent (SGD) can be applied on mini-batches. 
    
    In practice, it is hard to ensure the \textit{identically distributed} assumption as we cannot directly access the probability distribution $F(z)$, which is fixed but unknown. Instead, we have to collect data based on domain experiences and try to cover all possible cases. The general principle is that the collected data should follow the underlying probability distribution $F(z)$. Otherwise, the learned function will lead to many failure cases where data are poorly covered. This also explains why it is so difficult to develop a machine learning model for a general domain. Instead, one has to narrow down the target domain and try to close the loop of data collection and model development in real applications.
    \item \textbf{Limited function capacity} -- This assumption matches with our daily experience. For example, Occam's razor principle states that, of two explanations that account for all the facts, the simpler one is more likely to be correct. The statistical learning field has developed many theories to measure the capacity of a given set of functions $Q(z, \alpha)$, $\alpha\in\Lambda$. VC-dimension is one of such theories that has guided the development of the support vector machine (SVM) algorithm. However, it remains unclear how to explain deep learning models, which have demonstrated a remarkable generalization capability in many applications, yet seem to show an unlimited capacity in fitting any randomly labeled training data \citep{zhang2017understanding}. One of the most commonly used tricks in deep learning training approaches is data augmentation which ensures local smoothness of the learned function as adding small amount of noises to any input sample does not change its target label. The local smoothness property to some extend limits the capacity of the learning function as the function cannot fit arbitrarily changed labels in a local neighborhood of any input sample. Some other research \citep{nagarajan2019uniform} shows that this problem might be coupled with the SGD optimization process, which acts as an implicit regularization but is not included in the loss function. The deep learning generalization problem remains a large space for future studies.
\end{itemize}

\section{The Impact of Statistical Foundation on Computer Vision}
Computer vision can be generally divided into two major areas: geometry understanding and content understanding. Geometry understanding is largely driven by 3D reconstruction with projective geometry as its mathematical foundation, whereas content understanding is mostly concerned with visual recognition with statistical learning as its mathematical foundation.

This section mainly discusses computer vision problems related to visual recognition. We will see successful cases are often constrained to relatively narrow domains and where data can be easily collected and scaled up.

\subsection{Successful Cases in Visual Recognition}
Face detection is probably the first solved recognition problem in computer vision. The first breakthrough was taken place in 2001, thanks to the Viola–Jones object detection framework \citep{viola2001rapid}. In early 2000s, the training data for face detection is from a few thousands to tens of thousands. But it already led to many real applications. The most representative one is the hardware-enabled face detection module widely used in digital cameras for improving face-prioritized focusing function~\citep{dpreview2005nikon}. Since 2012, the research community has developed much larger datasets for face detection, e.g. 367,888 face annotations for 8,277 subjects in the UMDFaces dataset~\citep{bansal2017umdfaces} and 393,703 faces from 32,203 images with a high degree of variability in scale, pose and occlusion in the WIDER Face dataset~\citep{yang2016wider}. As a result, the face detection performance has been significantly improved from AP 58.8\% in 2014 to 95.7\% in 2019 as measured on the WIDER Face medium set~\citep{paperwithcode2022wider}.

Face recognition has lagged behind face detection for many years but started to blossom since 2012 due to three factors. First, the breakthrough in deep learning algorithms made end-to-end feature learning possible and effective. Second, large-scale face recognition data sets developed in the research community, including CASIA WebFace~\citep{yi2014learning}, MS-Celeb-1M~\citep{guo2016ms}, VGG-Face2~\citep{cao2017vggface2}, etc, greatly improved the robustness and generalization ability of the learned representation and motivated more researches on new loss functions that encourage large margin and better generalization. Third, the deployment in real applications further closed the loop in scenario-targeted data collection and effectively increased the coverage of tail data that are under sampled in general face recognition data sets. However, while face recognition has been able to leverage hundreds of millions of face images, it still suffers from the bias issue and is vulnerable to adversarial attack, indicating the extreme difficulty of satisfying the asymptotic uniform convergence condition which requires sufficiently large number of samples. Nevertheless, from the research perspective, face recognition is a unique domain that can setup a classification problem with millions of mutually exclusive classes. Moreover, face images are normally cropped and aligned, leading to a data manifold of much lower dimension. This makes face recognition an ideal problem for representation learning study.

In other domains, despite great progress in the past few years, there is a big gap between the SOTA and the desired performance in real applications. The progress is mainly on the algorithm part, i.e. the mapping function $f(x|\alpha)$ in terms of the statistical learning formulation as in Eq. \eqref{eq:loss_expected}. However, the progress is greatly limited by the lack of sufficient data, regardless of image classification, object detection, object/scene segmentation, or motion-related video analysis. The more general a visual recognition problem is, the harder it is to address the data shortage. To address this challenge, a typical practice is to limit the problem to a narrow domain to reduce the variety of the input data space, letting the underlying probability distribution $F(x,y)$ concentrate on a small region and thus making it easier to fulfill the i.i.d sample assumption. Another practice is to collect more real data after a model is deployed, building a closed loop for data collection and model improvement. This has been a practical paradigm in many applications.

While such practices have led to many successful vision applications, they also led to many issues such as scattered model development efforts, hard-to-maintain pipelines, limited model generalization capability, and being vulnerable to attacks, etc.

\subsection{Towards Generic Representation Learning}
To address the aforementioned issues, recent years have witnessed great progress in visual representation learning, with the goal to leverage a much larger scale of weakly labeled data to learn better visual representations via pre-training. For example, \citet{kolesnikov2020big} scaled up pre-training on the JFT-300M dataset, which contains 300M noisily labeled images. Their largest pre-trained model BiT-L shows remarkable transferring capabilities after being fine-tuned on a wide range of data sets. For example, BiT achieves 87.5\% top-1 accuracy on ImageNet and 76.3\% on the 19 task Visual Task Adaptation Benchmark (VTAB). \citet{dosovitskiy2020image} for the first time introduced Transformers to replace convolutional networks for image classification. When pre-trained on the same JFT-300M dataset, Vision Transformers (ViT) reaches the accuracy of 88.55\% on ImageNet. This work not only established a new SOTA on ImageNet, but also motivated many model architecture improvements by introducing better inductive biases to compensate Transformers which only rely on global attention. A natural idea is to marry convolutions or local attentions with Transformers to enjoy both local structure (and thus translation-invariant) and long range attention. For example, CvT~\citep{wu2021cvt} introduces convolution projections to enhance the linear projections for query, key, and value tokens. Swin Transformer~\citep{liu2021swin} proposes to use a shifted windowing scheme by limiting self-attention computation to non-overlapping local windows for greater efficiency. CoAtNet~\citep{dai2021coatnet} develops a family of hybrid models to stack convolution layers and Transformer layers. By pre-training on an enlarged JFT-3B dataset, CoAtNet further refreshed the SOTA on ImageNet with an accuracy of 90.88\%. 

The above works mainly train models using weakly labeled classification data, i.e. images with noisy keywords, and the learned representation can only be transferred to classification tasks. In contrast, \citet{radford2020learning} developed a contrastive language–image pre-training (CLIP) approach which can leverage a wide variety of images with noisy natural language supervision that is abundantly available on the internet. By pre-training on 400M images with noisy language supervision, the learned visual representation is well aligned with language representation and achieves competitive zero-shot performance on a great variety of image classification datasets. For example, CLIP ViT-L achieves a zero-shot image classification accuracy of 76.2\% on ImageNet and matches the performance of ResNet-50 without using any training images from ImageNet. \citet{jia2021scaling} scaled up the training data to 1.8B image alt-text pairs collected on the web to train a model named ALIGN (A Large-scale ImaGe and Noisy-text embedding). The aligned visual and language representation further improves zero-shot image classification to 76.4\% and also sets new SOTA results on Flickr30K~\citep{plummer2015flickr30k} and MSCOCO~\citep{lin2014microsoft} image-text retrieval benchmarks. \citet{yu2022coca} developed an image-text encoder-decoder foundation model called CoCa (Contrastive Captioner), pre-trained the model on JFT-3B (3B weakly labeled images) and ALIGN (1.8B image-text pairs), and achieved a new SOTA on ImageNet with an accuracy of 91.0\%.

CLIP marks the beginning of large-scale vision-language multi-modality representation learning. However, the visual representation learned in both CLIP and ALIGN are at the image level, which can be mainly used for image classification and cross-modal retrieval tasks. To expand the representations to a wider scope of vision tasks, \citet{yuan2021florence} developed a new vision foundation model called Florence, which is trained on 900M images with 900M free-form texts. To bridge the gap between the generic pre-trained representation and the final downstream tasks of great varieties, such as object detection, vision-language tasks (e.g. image captioning and visual question answering), and video understanding, Florence introduces several adapters which further fine-tune the generic representation for different task domains. This foundation model achieves new SOTA results on a majority of 44 representative benchmarks, e.g. ImageNet-1K zero-shot classification with top-1 accuracy of 83.74\%, 62.4\% mAP on COCO detection, 80.36\% on VQA, and 87.8\% on Kinetics-600.

The progress towards generic representation learning is closely coupled with the increase of training data, which are typically of billion images nowadays. Such a trend can be regarded as an application of the uniform convergence as shown in Eq. \eqref{eq:uniform_convergence}. The goal of generic representation learning should be to minimize the expected loss as in Eq. \eqref{eq:loss_expected}, so that the learned representation works the best for all possible visual inputs. However, due to many practical limitations, such as non-trivial and even biased training data collection, limited computing resource, and imperfect mapping function (model architecture), we can only perform representation learning via Eq. \eqref{eq:loss_emp}. As the uniform convergence is concerned with asymptotic convergence, it only shows us a trend of continued performance improvement with more training data used. To understand such a trend, \citet{kaplan2020scaling} and \citet{zhai2021scaling} studied the empirical scaling laws of neural language models and disciminative image models, respectively. Both papers find power laws that can describe the relationships between compute, data size, model size, and performance. While billions of images have presented tremendous technical challenges, considering the nature of the complexity of our visual world, we believe the trend of scaling up training data will continue. However, it does not mean that adding more data is the only way to improve the generalization of the learned representation. Introducing better inductive biases to capture intrinsic structures of visual concepts can make the learning more data-efficient.    

\section{Discussions}\label{sec:discussion}
We have revisited the statistical foundation behind machine learning and how it impacts visual representation learning. With the development of deep learning, the computer vision community has made remarkable progress in the past decade by scaling up data, model, and compute. In this section, we discuss several questions related to visual representation and multimodal learning which are frequently asked and discussed in the public media and research community. We attempt to answer and discuss them from the perspective of statistical learning.

\begin{question}
Are giant models really necessary?
\end{question}

In the past few years, we have seen great process in developing gigantic deep learning models for both language understanding and visual recognition. Such giant models typically have billions or even trillions of parameters and are also called foundation models~\citep{bommasani2021opportunities} for their great adaptability to a wide range of downstream tasks. For example,
BiT~\citep{kolesnikov2020big}, ViT~\citep{dosovitskiy2020image}, CLIP~\citep{radford2020learning}, ALIGN~\citep{jia2021scaling}, SimVLM~\citep{wang2021simvlm}, CoCa~\citep{yu2022coca}, Florence~\citep{yuan2021florence}, and GIT~\citep{wang2022git} as vision or vision-language models have led to SOTA's on many downstream tasks, while they also received many critiques for their limited capability in reasoning. 

If we view the progress from the perspective of statistical learning, Eq. \eqref{eq:uniform_convergence} shows that big data is a necessary condition to ensure the convergence of an empirical loss to its expected loss. See our discussion about big data in Sec. \ref{sec:foundation}. Even for a simple face/non-face classification problem for an input image of $20\times20$ with 16-level gray values, it is impossible to enumerate all possible images for training using our current computing power, let alone more challenging problems such as generic language understanding and visual recognition. To gain a better performance using statistical learning-based approaches such as deep learning, scaling up data is a necessary condition. Given the complexity brought in by super large-scale data, giant models are highly desirable for their greater capacities. 

In terms of uniform convergence, the SOTA scale of the training data used for training language or vision models is still far from sufficiently large. On one hand, we foresee continued progress of giant models in the next few years. On the other hand, for open domain problems such as language understanding and generic visual recognition, it is impossible to solve the problem merely by scaling up data and model. New breakthroughs from algorithms to approaches need to be investigated and explored.

\begin{question}
\label{question:unsupervised}
Why is unsupervised learning so critical for representation learning?
\end{question}

In Sec. \ref{subsec:loss}, we have shown several loss functions $Q(f(x|\alpha),y)$ for supervised learning to measure the difference between the predicted value $f(x|\alpha)$ for a given input $x$ and its corresponding ground truth target $y$. We also briefly mention that the loss function $Q(f(x|\alpha),y)$ also applies to unsupervised learning, where the supervision $y$ is simply the input $x$ itself, i.e. $y=x$. Below is an example for L2-based loss, while many other forms can also be introduced.
$$
Q(f(x|\alpha),x)=\norm{x-f(x|\alpha))}_2^2.
$$

Such a loss is essentially to perform a reconstruction task, with interest to discover intrinsic structures from data for more robust representation learning, e.g. finding low-rank structures on data manifold (sparse coding, auto-encoder, K-means, principle component analysis, Gaussian mixture model, etc.), disentangling multiple factors that control data generation (decoupling shape and appearance representations), masking and reconstructing tokens for contextualized representation learning, and predicting next tokens in an auto-regressive way for sequence generation.

Due to the strong desire of using large-scale data to ensure (or approach to) the asymptotic convergence to the true underlining expected training loss, unsupervised learning has become a fundamental approach to representation learning as it can easily leverage unlimited amount of training data without additional annotations. 

Below we list several practical examples recently used for either language, vision, or language-vision multi-modality representation learning, and how they are connected with unsupervised learning:
\begin{itemize}
    \item \textbf{Text to text}: This is the most widely used form in natural language processing, where corrupted  human-written text is used as input to predict (or reconstruct) the corrupted words. The corruption could be either random token mask or left-to-right auto regression. The recent SOTA language models such as BERT~\citep{devlin2019bert}, GPT-3~\citep{brown2020language}, Switch Transformer~\citep{fedus2021switch}, and Wudao 2.0~\citep{wudao2021} are all trained in this way.
	\item \textbf{Image to image}: This is related to many classic computer vision problems, such as low-level image processing (super-resolution, denoising, deraining, etc.), face synthesis, style generation, and image auto regression. Image Processing Transformer (IPT)~\citep{chen2021pre} and Image GPT~\citep{chen2020generative} are two early examples demonstrating the power of Transformers for image reconstruction. BEiT~\citep{bao2021beit, peng2022beit, wang2022image}, Masked Autoencoder (MAE)~\citep{he2021masked}, Contextual Autoencoder (CAE)~\citep{chen2022context}, and Masked Image Modeling (MIM)~\citep{xie2022simmim} are other examples for visual representation learning by masking random patches of an input image and reconstructing the missing pixels.
	\item \textbf{Image to text}: Assuming the input includes paired image and text, this is to generate image caption in an auto regressive way. For example, the Oscar pre-training~\citep{li2020oscar} is designed similarly as in text-to-text training, except that it has additional visual tokens extracted from images. Likewise, both CoCa~\citep{yu2022coca} and GIT~\citep{wang2022git} include an auto regressive caption generation process conditioned on an input image. 
	\item \textbf{Text to image}: While this is not a new topic, its performance is far from satisfactory due to the limited information conveyed in an input text and a large degree of freedom in the image space. Recently, DALL.E~\citep{ramesh2021zero}, GLIDE~\citep{nichol2021glide}, DALL.E 2~\citep{ramesh2022hierarchical}, Imagen~\citep{saharia2022photorealistic} show a new potential of text-to-image generation with impressive generation results. It again confirms the necessity of using large-scale data for solving such a challenging problem.
\end{itemize}

While reconstruction-based learning is a natural choice for unsupervised learning, other self-supervised and weakly-supervised learning are also indispensable when large-scale datasets with either intrinsic structures or weak labels are available. For example:
\begin{itemize}
    \item \textbf{Self-supervised learning}: MoCo~\citep{he2020momentum} and SimCLR~\citep{chen2020simple} leverage image intrinsic structures and perform contrastive learning to learn representations by maximizing agreement between differently augmented views of the same image.
	
	\item \textbf{Contrastive Language-Image Pre-training}: CLIP~\citep{radford2020learning} leverages 400M (image, text) pairs collected from the internet to perform contrastive learning to learn representations by maximizing agreement between paired image and text.
\end{itemize}

\begin{question}
Why is pre-training needed?
\end{question}

Model pre-training and fine-tuning has been a widely used practice in computer vision. For example, object detection normally pre-trains classification-related model parameters on ImageNet and then fine-tune all parameters on a detection data set~\citep{ren2015faster}.

Similar to computer vision, NLP has also seen great progress in the areas where large-scale data sets have been developed, for examples, machine translation, but falls short when it is hard to collect substantial amount of labeled data. To address the data challenge, the NLP research community has shifted from the problem-specific model development mode to the paradigm of pre-training and fine-tuning, as evidenced by BERT~\citep{devlin2019bert} and GPT-3~\citep{brown2020language}. The phenomenal success of model pre-training in NLP has greatly motivated computer vision researchers to explore the new paradigm of pre-training and fine-tuning, e.g. ViT~\citep{dosovitskiy2020image}, CLIP~\citep{radford2020learning}, Florence~\citep{yuan2021florence}, CoCa~\citep{yu2022coca}, etc. 
 
Despite many differences between computer vision and NLP, if we take a step back and think about their similarities in terms of statistical learning, we can learn many things from each other.

For instance, the spirit of pre-training is that one can leverage unlimited amount of data without labels (or with weak labels) to learn features that conform with downstream tasks. Then with very limited (e.g. few-shot) supervision, the learned feature can lead to a high accuracy on downstream tasks. This makes the paradigm of pre-training and fine-tuning an effective solution to solving (or mitigating) the data shortage problem.

The connection between pre-training and fine-tuning can be seen in Table \ref{table:connection}. 

\begin{table*}[ht]
\small
\newcolumntype{M}[1]{>{\centering\arraybackslash}m{#1}}
{\renewcommand{\arraystretch}{1.5}
\caption{Connection between pre-training and fine-tuning}
\label{table:connection}
 \begin{tabular}{|m{0.14\linewidth}|m{0.38\linewidth}|m{0.38\linewidth}|} 
 \hline
  & \multicolumn{1}{c|}{\textbf{Pre-training}} & \multicolumn{1}{c|}{\textbf{Fine-tuning}} \\  
 \hline
    Overall \newline training loss  &
    $$L_{emp}(\alpha)=\frac{1}{N}\sum_{i=1}^{N}Q(f(x_i|\alpha),y_i)$$ 
    $N$ is normally much larger.
    
    &
    $$L'_{emp}(\alpha)=\frac{1}{N}\sum_{i=1}^{M}Q'(f(x_i|\alpha),y_i)$$ 
    $M$ is small, $M\ll N$.
    
    \\
 \hline
 Typical \newline training tasks & 
 \begin{itemize}[leftmargin=.1in]
     \item 	\textbf{Unsupervised}: Reconstruction-based loss, where we do not have supervised label $y_i$ but can use $x_i$ as a self-supervision target:
     $$Q(x,f(x|\alpha))=\norm{x-f(x|\alpha))}_2^2$$
     $KL$ distance can also be used when $x$ can be converted to a probability. E.g. auto regression or masked token loss for self-supervised image pre-training (BEiT).
     \item \textbf{Self-supervised}: Contrastive learning for image multi-crops, e.g. MoCo and SimCLR.
     \item \textbf{Weakly-supervised}: Contrastive learning for multi-modality data, e.g. CLIP.
 \end{itemize} &
  \begin{itemize}[leftmargin=.1in]
     \item \textbf{Supervised}:
     $$Q'(y,f(x|\alpha))=\norm{y-f(x|\alpha))}_2^2$$
     \newline \newline \newline \newline \newline \newline \newline \newline \newline \newline \newline \newline 
 \end{itemize}\\[2cm]
 \hline
 Data set scale & 
 Training tasks are designed to easily scale up data for better convergence of the empirical loss to the expected loss. &
 Small-scale (even few-shot) high-precision supervised data for model adaptation. \\
 \hline
 Connection & 
 \multicolumn{2}{m{0.8\linewidth}|}{The connection is the shared function $f(x|\alpha)$, which can be slightly fine-tuned or directly used in downstream tasks, thanks to the layered deep function design, making feature reuse more effective.} \\
 \hline
\end{tabular}
}
\end{table*}

In the table, we use examples mainly from vision. We want to particularly echo the importance of reconstruction-based loss (see the related discussion in Question \ref{question:unsupervised}) for its unsupervised nature, making it possible to utilize unlimited amount of data without additional manual annotations. 

Table \ref{table:connection} not only provides a statistical justification of the pre-training and fine-tuning paradigm, but also points out the significance of big data for generic representation learning. 

The connection between pre-training and fine-tuning is the shared function $f(x|\alpha)$, which can be largely reused (generalized) in downstream tasks with slight fine-tuning on small-scale high-precision supervised data. The generalization ability mainly comes from two ways. The first is that the large-scale data set used in pre-training helps to define a better approximation to the expected loss. The second is that the layer-by-layer designed deep learning function includes hierarchical and multiple explanatory factors to allow re-using parameters in a combinatorially efficient way \citep{bengio2013representation}. As both pre-training and fine-tuning use stochastic gradient decent (SGD) approaches to search for optimal parameters, pre-training can also be interpreted as finding and providing a better initialization location in the parameter space for fine-tuning optimization. A recent research work \citep{nagarajan2019uniform} shows that a larger-scale training set can let SGD explore a larger parameter space, leading to a better parameter that cannot be found by only training on a smaller-scale training set. However, it remains an open problem to fully understand the generalization ability of the learned representation through pre-training.

This new paradigm of large-scale pre-training naturally leads us to rethink vision problems in data construction, learning algorithm design, and scalable infrastructure development.

\begin{question}
Is it possible for machine learning to generalize?
\end{question}
Generalization is a widely but often vaguely used term in computer vision and natural language processing. As the goal of machine learning is to make predictions for unseen samples, a common practice is to use a test set or multiple test sets, as a surrogate to the expected loss, to measure the prediction accuracy of the learned model. In many cases, data-efficient learning algorithms are highly demanded so that one can use less amount of training data to achieve better prediction accuracy, which is often referred to as better generalization. To improve the generalization ability of a learning algorithm, many general principles have been developed in the past, such as L2 regularization~\citep{tikhonov1995numerical} and large margin~\citep{vapnik1998}, which help avoid over-fitting by adding very general prior information to regularize the learning function $f(x|\alpha)$.

While great progress has been made in developing general learning theories and practical algorithms, one often compares machine learning with human intelligence, hoping to gain even better ``generalization'' abilities, such as commonsense, interpretability, and reasoning capabilities. It is worth noting that statistical learning alone cannot lead to such better ``generalizations.'' Uniform convergence only tells us the sufficient and necessary conditions of the empirical loss converging to the expected loss. Even when the convergence happens, the best mapping function $f(x|\alpha^*)$ parameterized by $\alpha^*$ that minimizes the expected loss cannot yield more desired ``generalization'' abilities. The root cause is due to the black box function used in statistical learning, or more recently in deep learning. 

As machine learning can be regarded as an application of the inductive reasoning principle, it would be interesting to revisit the development of Newton's law of universal gravitation, which is an unprecedented successful case of inductive reasoning. It demonstrates great ``generalization'' abilities, with the result interpretable and conforming with commonsense. The ``generalization'' abilities are the result of the white box approaches employed by Newton, including the use of Newton's laws of motion and the infinitesimal calculus developed by him, to explain planetary motion. Prior to him, Kepler also developed three laws of planetary motion by analyzing the astronomical observations about orbits of planets around the Sun. Both Kepler and Newton applied the inductive reasoning principle, to explain why planets move around the Sun in the way as observed by previous astronomers. But Kepler's approach is more data-driven, extracting scientific discoveries through the analysis of data, whereas Newton's approach is first-principle-based, aiming at discovering the fundamental principles that govern the world by applying the laws of motion and calculus~\citep{e2021dawning}. 
Note that both approaches are still explaining seen data for small masses or those travelling in slow speeds, but fail in generalizing to out-of-distribution data when the speed of an object approaches the speed of light.

Unlike pure data-driven approaches, Newton's approach is much more data-efficient, but leads to a strong generalization ability. This is largely due to the while box approaches used in the derivation of the law of universal gravitation. For example, the Newton's law of motion and the infinitesimal calculus introduce causal relationships between force and motion, which are the result of deductive reasoning and hold beyond the observed orbits of planets.

Similarly, the convolution operation introduces the translation-invariant property which does not need to be learned from data but generally holds for any 2D images. This operation effectively eliminates the need of redundant training data with same objects appearing at different locations. Likewise, Transformers utilize the attention mechanism to fuse information over multiple tokens, which effectively capture the intrinsic structure of language. After Transformers were introduced to vision, researchers have found it very beneficial to introduce hierarchical structure (for multi-scale features)~\citep{wang2021pvt}, constrain global attention (for local spatial modeling)~\citep{liu2021swin}, and marry attentions with convolutions~\citep{wu2021cvt}. Such structures built into deep neural networks are normally called inductive biases, which were believed at the heart of the ability to generalize beyond observed data~\citep{mitchell1980need}. Future efforts to study machine learning must focus on adding structures and prior knowledge to the learning process, making the learned representations capable of performing reasoning, interpreting predictions, and thus generalizing beyond observed data. Recently, there have been a few attempts to introduce part-whole structures~\citep{hinton2021represent}, to develop a cognitive architecture with a world model that can be trained in a self-supervised manner~\citep{lecun2022path}, or to impose the principles of parsimony and self-consistency for a white-box and closed-loop learning~\citep{ma2022principles}.

\section{Future Trends}
We are seeing a transformation in both vision and language areas to pursue large-scale pre-training for representation learning, as evidenced by the explosive growth of big model training in NLP and vision. This coincides with the statistical foundation behind machine learning. The use of large-scale data leads to a better convergence of an empirical loss to its expected loss. As a result, the learned representation is more robust and generalizes better in downstream tasks. 

We discuss in this section several future trends which call for both system-level and algorithm-level researches.  

\textbf{Continued increase of training data}: Data is never sufficient in real computer vision applications. As long as a learning problem is formulated properly, adding data will always improve the convergence of an empirically learned model to its expected model. As it is generally hard to scale up supervised data that require manual annotations, the increase of data will be eventually unsupervised or weakly supervised, which can be collected from the web, such as text, images, videos, and multi-modality data. And recent progress in new model architectures and learning approaches such as Transformer and contrastive learning makes it possible to consume super large-scale training data with noises. Proprietary data collected from enterprise scenarios will further build a competing advantage in related applications. 

There is, however, a question of diminishing return when more and more data is added to help train a bigger and bigger model. In practice, it is often the small amount of data for a specific domain that makes the real difference in system performance. 

\textbf{System-level optimization}: The use of web-scale data significantly increases the computational cost, let alone its demand of larger model requires more computational budget. As a result, it is not uncommon that a pre-training task uses hundreds or thousands of modern GPUs. This introduces many system-level problems, such as efficient communication between compute nodes, data partition and model partition across GPUs, memory-efficient optimizer allowing to use larger batch size, etc. Large-scale gradient decent-based optimization also leads to many numerical and training convergence issues, such as overflow and underflow in float point representation especially when half-precision like FP16 is used. Building a scalable training infrastructure with system-level optimization will be indispensable for large-scale training.

\textbf{Novel pre-training tasks}: Compared with NLP, vision problems are more diverse, varying by their prediction granularity: classification, object detection, segmentation, pose estimation, motion and tracking, image/video captioning, etc. each requiring different solutions. Novel pre-training tasks for vision that can benefit multiple problems, leverage larger-scale training data, and boost downstream vision problems will be appreciated. We also foresee that multi-modality learning for both image and video will become a mainstream direction for its great  potential of combining supervisions from multi-modality signals and learning grounded representation to improve both vision and language problems.

\textbf{Synthesis and simulation}: As data is never sufficient in terms of asymptotic uniform convergence, a large-scale simulation system that can generate close-to-real images of scenes with diverse view, lighting, and texture will be of great value to compensate the lack of data. Some problem domains, such as autonomous driving and human analysis, have seen very encouraging progress of training data generated by simulation. We expect continued progress in more domains where computer graphics techniques are improving as driven by strong industrial needs in game, digital twin, and metaverse. The synthetic data might be particularly more helpful when combined with pre-training tasks which can leverage both real and synthetic data and are more resilient to noise. Moreover, synthesis, combined with analysis, which is called analysis by synthesis, is also a crucial way to testing the analysis result, performing disentangled learning, and making analysis more explainable. Developing algorithms that can combine analysis and synthesis in a large-scale system will become more mainstream.

\textbf{Adding structure and knowledge}: With more data used, the learned representation will be statistically more robust and expressive. This provides a solid foundation for adding structure and knowledge to the learned representation and pursuing robustness, interpretability, commonsense, and reasoning capabilities in advancing artificial intelligence. An example is the recent image generation work DALL.E, which first trains a VQ-VAE model on large-scale unlabeled images (essentially a reconstruction task) to learn a compact representation and then combines the learned representation with language representation to perform conditional generation. Such kind of unsupervised and reconstruction-based learning will be a promising way for introducing structure and knowledge, disentangling multiple factors, and adding human priors into representations. This is made possible by the mathematically guaranteed uniform convergence behind data-driven machine learning.

\bibliography{bibliography}

\end{document}